\documentclass[a4paper]{article}
\usepackage{INTERSPEECH2019,amsmath,graphicx}
\usepackage{tipa}
\usepackage{mathtools}

\title{Neural Machine Translation for Multilingual Grapheme-to-Phoneme Conversion}
\name{Alex Sokolov, Tracy Rohlin, Ariya Rastrow}
\address{Amazon.com, Inc. USA}
\email{{alexsoko,tracyroh,arastrow}@amazon.com}

\begin{document}
\setlength{\parindent}{3ex}
\topmargin=0mm
%
\maketitle
\begin{abstract}

Grapheme-to-phoneme (G2P) models are a key component in Automatic Speech Recognition (ASR) systems, such as the ASR system in Alexa, as they are used to generate pronunciations for out-of-vocabulary words that do not exist in the pronunciation lexicons (mappings like $"e\ c\ h\ o" \rightarrow "E\ k\ oU"$). 

Most G2P systems are monolingual and based on traditional joint-sequence based n-gram models \cite{Novak2012WFSTBasedGC,journals/speech/BisaniN08}. As an alternative, we present a single end-to-end trained neural G2P model that shares same encoder and decoder across multiple languages. This allows the model to utilize a combination of universal symbol inventories of Latin-like alphabets and cross-linguistically shared feature representations. Such model is especially useful in the scenarios of low resource languages and code switching/foreign words, where the pronunciations in one language need to be adapted to other locales or accents. We further experiment with word language distribution vector as an additional training target in order to improve system performance by helping the model decouple pronunciations across a variety of languages in the parameter space. We show 7.2\% average improvement in phoneme error rate over low resource languages and no degradation over high resource ones compared to monolingual baselines.
\end{abstract}


\section{Introduction}
\label{sec:intro}
Alexa's ASR platform relies on a large hand-curated lexicon which is comprised of word-pronunciation pairs. This lexicon can never provide complete coverage over the vocabulary as it is often not worth the time or cost required to create these phonetic sequence mappings compared to how infrequently certain words occur during any given Alexa interaction. 

Grapheme to Phoneme systems, on the other hand, can learn these mappings automatically with high accuracy and are responsible for transcribing any out-of-vocabulary (OOV) tokens into phonemic representations.  These phonemic representations are an important component that lies between the language model and acoustic model of an ASR system. These OOV tokens often include rare and foreign words, the amount of which may vary depending on the language.

The challenge in designing the G2P system is to create a many-to-many mapping system that will learn not only the mapping between one grapheme and one phoneme, but also where one phoneme is represented by multiple graphemes (such as $"s\ h" \rightarrow$ "\textesh"). And for certain languages like English, these mappings can be inconsistent and ambiguous, especially in the case of names and foreign words.

Sequence to sequence (Seq2Seq) neural network models are one such way of learning the mappings between graphemes and phonemes where the input and output sequence can vary in length. Originally designed for machine translation, they have been applied on wide variety of problems, such as generative language models.  More recent Seq2Seq models heavily incorporate attention mechanism \cite{luong2015effective} and residual learning, while other models use encoder-decoder architectures that are recurrent \cite{bahdanau2014neural, wu2016google,DBLP:journals/corr/JohnsonSLKWCTVW16,journals/corr/Pouget-AbadieBMCB14,DBLP:journals/corr/GehringAGYD17} or self-attention \cite{vaswani2017attention} based. 

Recurrent seq2seq models pose a distinct advantage due to their ability to take in the input history when determining the output state, often outperforming n-gram models on classification tasks due to an n-gram’s heavy dependence on the previous n graphemes \cite{Mikolov2011RNNLMRecurrentNN}. This means that Recurrent Neural Networks (RNN) are better suited for sequence problems where longer term context and “soft” input embeddings are important. Long-Short-Term Memory (LSTM) networks \cite{Hochreiter1997LongSM} are in turn better at handling longer sequences and can have more layer depth as they are less prone to diminishing and exploding gradients. Bi-directional LSTMs that consider both past and future contexts have become increasingly popular over RNNs or uni-directional LSTMs that only consider past contexts. 

Seq2Seq models can be trained on multiple languages at once and used in multi-task and multi-modal learning scenarios. In context of G2P conversion, they allow joint learning of the alignment and translation of graphemes and phonemes in an end-to-end fashion. Therefore, they are a natural fit for our multilingual G2P task, especially since we can train it on large pronunciation lexicons with relatively short sequences lengths.  Seq2Seq models have been found to perform better on these short sequences than compared to very long sequences  \cite{journals/corr/Pouget-AbadieBMCB14}. However, until now neural G2P models have not shown superior results on their own \cite{Yao2015SequencetoSequenceNN}, compared to traditional joint-sequence based n-gram models for G2P.

In this paper, we examine whether Seq2Seq LSTM models perform better compared to traditional joint-sequence based n-gram models, given the latest advancements in Seq2Seq modeling for single language pairs. In addition, we investigate whether we can build a single multilingual G2P model which may outperform individual models trained on single language lexicons by utilizing transfer learning thereby improving performance on under-resourced languages and foreign words from different locales. The goal of such a system is to have a single multilingual model that matches or improves over the results of monolingual models without the degradation in accuracy introduced by using multilingual dataset. In particular, the model must be able to distinguish between languages where the same grapheme is paired to different phones, such that the model does not learn a single pairing and apply this pairing to all instances of the grapheme, regardless of input language.  Thus, we wish to avoid situations where a larger lexicon overwhelms a smaller lexicon and erroneously labels a particular grapheme. By using Seq2Seq LSTM models we achieve better PER and WER than low-resource monolingual models, while reducing the potential influence larger-resource lexicons might have in a multilingual model. 

\section{Related work}
\label{sec:related}

Traditionally G2P is done with joint-sequence n-gram models \cite{Novak2012WFSTBasedGC,journals/speech/BisaniN08}, \cite{Novak2013FailureTF,Novak2016PhonetisaurusEG}. While these are still considered to be state of the art models best designed for single lexicons, they require explicit alignment information to be provided during training. 

The Seq2Seq approach to G2P was first proposed in \cite{Yao2015SequencetoSequenceNN}, but the work only showed improvement over state of the art results when using the same alignments in RNN slot tagger fashion. The alignments were generated by a separate HMM many-to-many alignment procedure \cite{Jiampojamarn2007ApplyingMA}, a weakness inherited from joint sequence models. \cite{Rao2015GraphemetophonemeCU} used connectionist temporal classification with output delays instead of Seq2Seq for joint alignment and translation which resulted in a significant improvement when compared to the Phonetisaurus \cite{Novak2012WFSTBasedGC} baseline implemented as finite-state transducer (FST). 

A Seq2Seq bi-directional LSTM were used in \cite{Behbahani2016PersianST} to perform G2P tasks on Persian, a particularly challenging task as often times vowels are not indicated in the language’s orthography and are often dependent on their positions within the sentence. \cite{Novak2012WFSTBasedGC} likewise uses an RNNLM to perform N-best re-scoring on the G2P alignments produced by their Weighted Finite State Machine, beating state-of-the-art models using CMUDict and NetTalk data.

Most related to this work, \cite{Milde2017MultitaskSM} trained a Seq2Seq G2P model with stacked bi-directional LSTMs using attention and residual connections on two languages at once, while \cite{Yolchuyeva_2019} used Transformer architecture \cite{vaswani2017attention} to perform the task. \cite{Peters2017MassivelyMN} used a similar model to train a G2P model on hundreds of low resource languages, some of them in zero-shot fashion as described in \cite{wu2016google}. The focus of this paper, however, is on achieving state of the art performance on a limited set of languages, both high and low resource, with a single multilingual model.


\section{Dataset}
\label{sec:dataset}

Amazon has an extensive English ASR lexicon containing word-pronunciation pairs for En-US and En-UK locales, in addition to a large German ASR lexicon. Other data sources we used included CMUDict and the Wiktionary dataset \cite{Schlippe2010WiktionaryAA} which includes lower-resource languages such as Czech, Finnish, Hungarian and Polish. The data was transcribed using the standard X-Sampa phone set and stripped of any syllabic or stress markers. Tokens that had rare grapheme or phone (less than 25 and 5 times, respectively) were considered to be noise and filtered out. The combined dataset contains 4 million grapheme-phoneme pairs from 18 languages (21.02\% of which comes from En-US data alone), 84 unique graphemes and 117 phones (including 9 phones found in en-UK dataset but not in en-US, comprised primarily of vowels). 

The majority of graphemes (90.8\%) are unique to one lexicon, but still have multiple phonemic representations within each lexicon, as only 22.79\% of graphemes had a single phonemic representation. Block sampling was used to ensure that consecutive words sorted in alphabetical order were placed in the same train, dev or test partition. This would not allow the models to "cheat" by seeing almost identical words during train and test time (i.e. a word $"dogs"$ pronounced as $"d\ O\ g\ z"$ for en-US and as $"d\ O\ k\ s"$ for de-DE).

\begin{figure*}[t]
     \centering
     \includegraphics[width=1\textwidth]{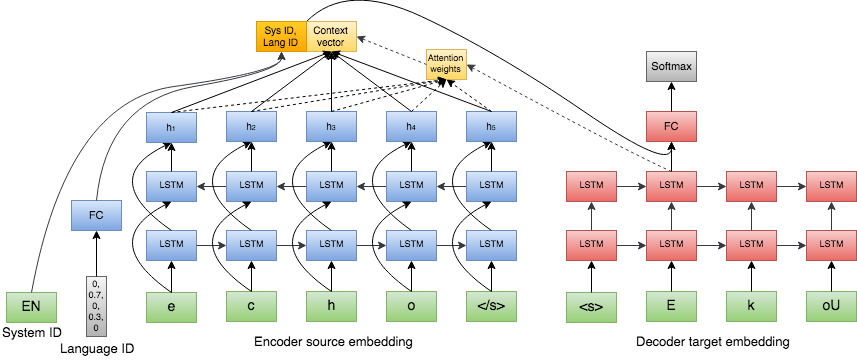}
     \caption{Proposed encoder-decoder model architecture for a single decoding step. Language distribution (Language-ID) and language label (System-ID) are concatenated together with attention context vector and fed to the decoder fully connected layer just before softmax.}
     \label{fig:model}
\end{figure*}

\section{Proposed model}
\label{sec:proposed}

The core of our model (Figure \ref{fig:model}) is an RNN encoder-decoder model with attention mechanism \cite{bahdanau2014neural}. The model consists of an encoder which compresses each source grapheme in the input sequence into a fixed-length vector, and a decoder which generates a phoneme sequence as output, conditioned on the attention over the encoder hidden states. They are trained jointly to minimize cross-entropy on the training data. For both the encoder and decoder, we use long short-term memory units (LSTM) \cite{Hochreiter1997LongSM}, which can process sequences of arbitrary length and use long histories efficiently. We use trainable character and phoneme embeddings as encoder and decoder inputs correspondingly.

The encoder is a stacked bidirectional LSTM, which processes the input in both forward and backward directions, as to represent both past and future dependencies at every time step. Forward and backward cells are stacked vertically and their outputs are concatenated at each layer.

The decoder is a language model conditioned on the past phoneme sequence and global attention mechanism \cite{luong2015effective}. Attention vectors are added to the decoder inputs which helps to facilitate information flow between the source and the target sequences. Instead of forcing the encoder to compress information about the whole input sequence in its last hidden state, a weighted sum over all encoder hidden states forms the context vector. At each decoding step, the decoder generates a softmax distribution over phonemes, until it outputs end-of-sentence symbol. N-best decoding is used to keep N different paths, while searching through the search space, to find more optimal solutions.

To generate pronunciation for the desired target language in a multilingual G2P scenario, we need to explicitly condition our model on the corresponding input parameter. Similar to \cite{DBLP:journals/corr/JohnsonSLKWCTVW16} we use a system ID (language token like $<$en-US$>$, $<$en-UK$>$, $<$de-DE$>$, $<$fr-FR$>$, etc) as input embedding, but instead of appending this token to the input sequence on the encoder side, we concatenate it with the attention context vector and feed into the decoder's fully-connected layer just before the softmax. The two approaches have similar performance, but our implementation is motivated by the need to have one other input parameter representing the language distribution (language ID) - a vector of length equal to the number of languages, which is represented by a set of tokens that are mutually exclusive to the tokens in the vocabulary. The language distribution vector represents the extent to which a particular word belongs to every language and is pre-computed using the following formula:
$$p(id \mid l)=\frac{C(w)}{log⁡\left | N \right |},\ \ \sum_{l}{p(id \mid l)}=1,$$
where $N$ refers to the lexicon size and $C(w)$ refers to the count of word occurrences found within a given language lexicon. We have also tried to smooth it by taking logarithm of word count and multi-hot binary vector representation, but that didn't have any significant impact on the results. The intuition  behind the language ID vector is that there may be a correlation between the word's origin language and its pronunciation for any given locale. Since it is hard for the model to learn language identification only having a single word as an input, the language ID might be a useful signal to help the model distinguish words having multiple pronunciations in different languages and model their phone distributions accordingly.
Since a very large number of words have several pronunciation alternatives even for the same language, we are doing n-best decoding during inference. We then select up to 3 best scoring hypothesis with average token posterior values above a threshold we optimized on the dev set (25\% for 2-best and 18\% for 3-best). This way, we increase the coverage of our output hypothesis without adding too much noise into it.

\section{Evaluation and results}
\label{sec:results}

We used Phoneme Error Rate (PER) and Word Error Rate (WER) metrics on the target phoneme sequence to evaluate the G2P models. PER is defined as $PER=\frac{LD(p,p')}{\left | p \right |}$, where $LD(p,p')$ represents the minimum (Levenshtein) distance between the predicted and ground truth phoneme sequence, length of which is represented by $\left | p \right |$. WER is defined as $WER=\frac{\left | E \right |}{\left | N \right |}$ , where $\left | E \right |$ represents the number of predicted pronunciations with one or more errors.

A baseline multilingual Seq2Seq model (System 0) which did not include language distribution and language label information performed very poorly across all the languages as expected and achieved average PER and WER of 10.09\% and 47.37\% , respectively. What was surprising, however is that System 2 also hasn't outperformed System 1 on any language, although it still performed much better than the baseline multilingual model, with a multilingual PER around 5.7\% and WER around 38.8\% for System 2. This might be due to the fact that language IDs are a function of the completeness of human labelled lexicons, sparsity of which might be introducing too much noise into the labels, especially for the languages for which we don't have a lot of annotations.

System 1 performed significantly better than the baseline model, with a 5.3\% PER and 36.4\% WER (table \ref{table1}). This must be due to the fact that by feeding the language label into the model, the model is receiving a “hint” as to which pronunciation is mapped to a particular grapheme. This helps to delineate between graphemes that might occur in multiple dialects but routinely have different phoneme (in particular with vowels) qualities. For example, the vowel "\textturnscripta" \ ($"Q"$ in X-Sampa notation) is much more likely to show up in British English than American English, as there are absolutely no instances of "\textturnscripta" in the en-US lexicon, but over 8,000 instances in the en-UK lexicon.

\begin{table}[!ht]
\centering
\caption{PER and WER results of baseline monolingual models and multilingual model + system ID (System 1) on large (\textgreater 300K words) and low-resource (\textless 15K words) lexicons. Best model (System 1) also didn't show any significant degradation compared to monolingual models for the high resource languages. Numbers in bold indicate languages for which the multilingual model improved in error rates, despite the class imbalance problem and opportunity for larger lexicons to influence the transcription.}
 \label{table1}
\begin{tabular}{|l|c|c|c|c|l|}
\hline
\multicolumn{5}{|c|}{\textbf{High resource languages}} &                            \\ \cline{1-5}
                      & \multicolumn{2}{c|}{\textbf{Monolingual}} & \multicolumn{2}{c|}{\textbf{Multilingual}} &                                               \\ \cline{1-5}
\textbf{Language}     & \textbf{PER}        & \textbf{WER}        & \textbf{PER}         & \textbf{WER}        &                                               \\ \cline{1-5}
\textbf{En-US}        & 0.127               & 0.525               & 0.129                & 0.528               &                                               \\ \cline{1-5}
\textbf{En-UK}        & 0.121               & 0.511               & 0.125                & 0.525               &                                               \\ \cline{1-5}
\textbf{German}       & 0.036               & 0.151               & 0.042                & 0.184               &                                               \\ \cline{1-5}
\textbf{French}       & 0.077               & 0.429               & 0.077                & 0.431               &                                               \\ \cline{1-5}
\textbf{Dutch}        & 0.023               & 0.121               & 0.028                & 0.135               &                                               \\ \cline{1-5}
\textbf{Italian}      & 0.003               & 0.027               & 0.005                & 0.046               &                                               \\ \cline{1-5}
\textbf{Hindi}        & \textbf{0.131}      & \textbf{0.584}      & \textbf{0.114}       & \textbf{0.545}      &                                               \\ \cline{1-5}
\multicolumn{5}{|c|}{\textbf{Low resource languages}}                                                          & \multicolumn{1}{c|}{\textbf{\# words}} \\ \hline
\textbf{Czech}        & \textbf{0.081}      & \textbf{0.489}      & \textbf{0.071}       & \textbf{0.439}      & \multicolumn{1}{c|}{5238}                     \\ \hline
\textbf{Finnish}      & \textbf{0.038}      & \textbf{0.322}      & \textbf{0.034}       & \textbf{0.288}      & \multicolumn{1}{c|}{8500}                     \\ \hline
\textbf{Hungarian}    & 0.014               & 0.143               & 0.015                & 0.155               & \multicolumn{1}{c|}{10132}                    \\ \hline
\textbf{Polish}       & 0.069               & 0.333               & 0.082                & 0.418               & \multicolumn{1}{c|}{7249}                     \\ \hline
\textbf{Portuguese}   & \textbf{0.149}      & \textbf{0.725}      & \textbf{0.130}       & \textbf{0.684}      & \multicolumn{1}{c|}{6567}                     \\ \hline
\textbf{Spanish}      & \textbf{0.094}      & \textbf{0.494}      & \textbf{0.053}       & \textbf{0.364}      & \multicolumn{1}{c|}{4916}                     \\ \hline
\end{tabular}
\end{table}

Certain languages also continue to perform better than others across all models. Italian, Dutch, and Hindi consistently have lower PER and WER in these models compared to the rest of the data set. It is important to note, however, that the Hindi dataset has graphemes transcribed into Latin-based characters and as such was likely phonetically transcribed. Still, the low PER and WER in these languages are likely due to a high one-to-one correspondence between graphemes and phonemes. This is in contrast to languages like English that require the model to learn many-to-many mappings (such as "th" to $"\theta"$). 
The many-to-many mapping is likely a factor in the high PER and WER for both dialects of English. In addition, certain languages in the dataset are more likely to use one particular grapheme for several different phonemes, leading to high PER and WER. Portuguese, for example, has several instances of one grapheme mapping to several phonemes, 
which is likely why this language consistently performs the worst out of any other language in the dataset.

In addition to these issues, certain datasets such as Portuguese and Polish likely suffered due to the mappings in these smaller datasets being out-weighed by larger lexicons like English and German. Additionally, the smaller datasets were often only comprised of Wiktionary phonetic transcriptions. \cite{Schlippe2010WiktionaryAA} discusses how the Wiktionary lexicons are consistently more likely to perform worse on G2P tasks compared to the researchers’ own GlobalPhone dataset, except for German. Since the Spanish, Portuguese, Polish, Finnish, Esperanto and Czech lexicons are comprised entirely of Wiktionary data, it is likely the quality of the original human-annotated transcriptions posted on Wiktionary that are causing high error rates among these lexicons.

\begin{figure}[!ht]
     \includegraphics[width=\linewidth]{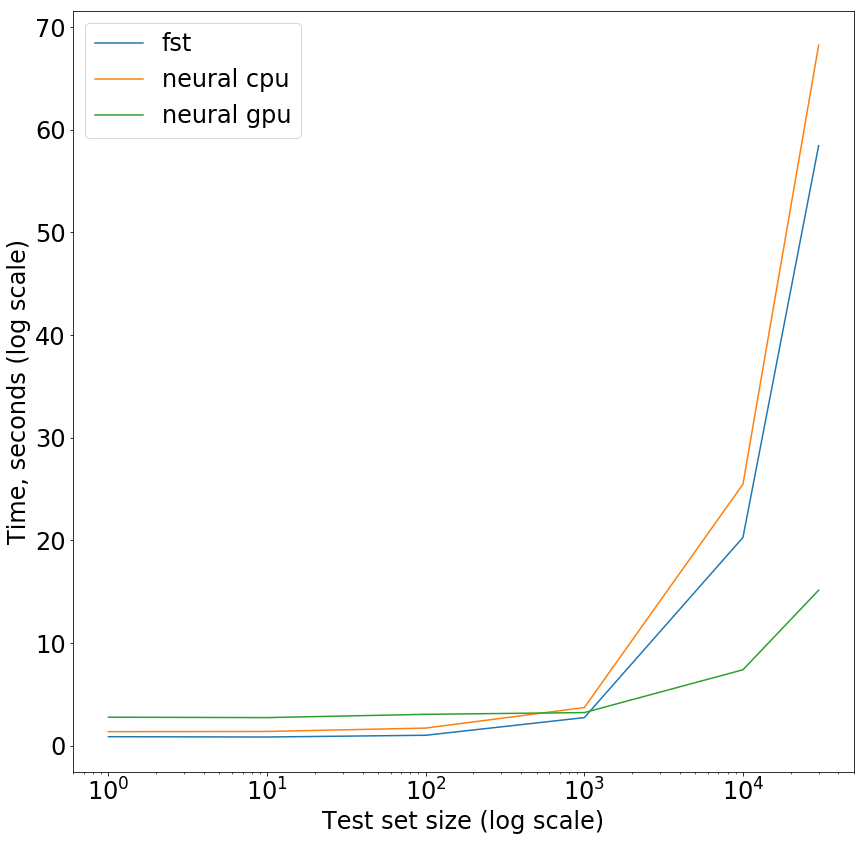}
     \caption{Inference latency of FST and neural G2P models for varying amounts of input data.}
     \label{fig:latency}
\end{figure}

Figure \ref{fig:latency} shows the inference latency of FST (Phonetisaurus on CPU) and proposed model (on CPU and GPU) in seconds. While fst is faster on CPU and for small amounts of input data, highly parallelized GPU based inference with sufficiently large batch size is much faster (7 times for batch size of 1024 on AWS Tesla V100 instance p3.2xlarge).

Model training time is very small – it takes about 0.25 hours to train a monolingual model on CMUDict dataset in Sockeye \cite{Hieber2017SockeyeAT} when using AWS p3.16xlarge, compared to 6 to 44 hours for previous systems reported in \cite{Milde2017MultitaskSM}. This greatly reduces the turn around time for hyper-parameter tuning when onboarding the model to new languages Alexa expands in.

\section{Conclusion and future work}
\label{sec:conclusion}

In this work, we propose a novel approach to G2P that allows us to exploit large amounts of multilingual data to enhance prediction accuracy compared to monolingual models, especially for low resource languages. Single multilingual G2P model is much more flexible and easier to maintain in production environment, compared to monolingual n-gram based models - it's easier to fine-tune it with the latest training data updates or adapt to a new language.

We are also experimenting with several general techniques, such as model ensembling, self-training, selecting n-best hypothesis based on the score of separately trained confidence model, optimizing on sequence level metric (phone error rate/phonemic distance) and using Transformers \cite{vaswani2017attention} as the core architecture. Each of these is bringing good accuracy improvements to the G2P model in general, although they are not directly related to the multilingual and low resource aspects we explored in this paper.





\bibliographystyle{IEEEtran}
\bibliography{strings,refs}

\end{document}